\documentclass{article}

\usepackage{microtype}
\usepackage{graphicx}
\usepackage{subfigure}
\usepackage{booktabs} 
\usepackage{tikz}
\usetikzlibrary{bayesnet}

\usepackage{hyperref}
\usepackage{amsmath}
\usepackage{amssymb}
\usepackage{pgfplots}
\newcommand{\meanone}{2}
\newcommand{\meantwo}{4}
\newcommand{\sigmaone}{0.5}
\newcommand{\sigmatwo}{0.5}
\newcommand{\sigmathree}{0.5}
\newcommand{\mixone}{0.8}
\newcommand{\domainstart}{0}
\newcommand{\domainend}{6}
\newcommand{\margin}{0.1}
\newcommand{\yshift}{1.8cm}
\newcommand{\meanheight}{2.4cm}
\newcommand{\mixheight}{1.4cm}
\newcommand{\mixheighttwo}{1.9cm}
\newcommand{\height}{4cm}
\newcommand{\width}{12cm}

\pgfmathdeclarefunction{gauss}{2}{%
  \pgfmathparse{1/(#2*sqrt(2*pi))*exp(-((x-#1)^2)/(2*#2^2))}%
}

\pgfmathdeclarefunction{mmgauss}{5}{%
  \pgfmathparse{1/(#2*sqrt(2*pi))*exp(-((x-#1)^2)/(2*#2^2)) * #5 + 1/( #4*sqrt(2*pi))*exp(-((x-#3)^2)/(2*#4^2)) * (1 - #5)}%
}
\pgfmathsetmacro{\meanthree}{(\meanone + \meantwo) / 2}
\pgfmathsetmacro{\mmdleft}{(\meanthree - \margin}
\pgfmathsetmacro{\mmdright}{(\meanthree + \margin}
\pgfmathsetmacro{\mileft}{(\meanone + 5 * \margin}
\pgfmathsetmacro{\miright}{(\meantwo - 5 * \margin}

\pgfmathsetmacro{\meanmixtureone}{\mixone * \meanone + (1-\mixone) * \meantwo }
\pgfmathsetmacro{\meanmixturetwo}{(1-\mixone) * \meanone + \mixone * \meantwo }
\pgfmathsetmacro{\meanmixturethree}{\mixone * \meanone + (1-\mixone) * \meanthree }

\usepackage[accepted]{icml2018}

\icmltitlerunning{Hierarchical VampPrior Variational Fair Auto-Encoder}

\begin{document}

\twocolumn[
\icmltitle{Hierarchical VampPrior Variational Fair Auto-Encoder}



\begin{icmlauthorlist}
\icmlauthor{Philip Botros}{UvA}
\icmlauthor{Jakub M. Tomczak}{UvA}
\end{icmlauthorlist}

\icmlaffiliation{UvA}{AMLAB, University of Amsterdam}
\icmlcorrespondingauthor{Philip Botros}{philip.botros@student.uva.nl}
\icmlcorrespondingauthor{Jakub M. Tomczak}{j.m.tomczak@uva.nl}

\icmlkeywords{Machine Learning, ICML}

\vskip 0.3in
]



\printAffiliationsAndNotice{} 

\begin{abstract}
Decision making is a process that is extremely prone to different biases. In this paper we consider learning fair representations that aim at removing nuisance (sensitive) information from the decision process. For this purpose, we propose to use deep generative modeling and adapt a hierarchical Variational Auto-Encoder to learn these fair representations. Moreover, we utilize the mutual information as a useful regularizer for enforcing fairness of a representation. In experiments on two benchmark datasets and two scenarios where the sensitive variables are fully and partially observable, we show that the proposed approach either outperforms or performs on par with the current best model.  
\end{abstract}

\section{Introduction}
Reducing bias in machine learning algorithms has been an active area of discussion recently after the reliance on algorithmic decision making has been greatly increased. Consider the case of credit assignment, mortgage approvals or the provision of health care, where there are growing concerns that biases based on historical data prevent a fair process. 

In these cases it is not sufficient to prevent the decision maker from having access to the sensitive variable since this information has already been leaked into other features \cite{dwork2012fairness, mcnamara2017provably, menon2017cost, zafar2017fairness}. To correct for this and ensure a fair process, a new representation has to be created where all the sensitive information is removed.


More formally, we can consider this problem as the task of learning fair representations, where the goal is to learn a representation $\mathbf{z}$ that maximizes the information about the class label $\mathbf{y}$, while removing the sensitive information $\mathbf{s}$.

Learning rich representations from vast amounts of data using deep generative models remains one of the major challenges of machine learning. In recent years, different approaches to achieving this goal were proposed by formulating alternative training objectives to the log-likelihood \cite{goodfellow2014generative} or by utilizing variational inference that leads to a highly scalable framework now known as the variational auto-encoders (VAE)  \cite{kingma2013auto, rezende2014stochastic}.

The use of a deep generative model for fair classification has already been explored by \cite{louizos2015variational} who proposed the Variational Fair Auto-Encoder (VFAE). They, however, do not consider the partially-supervised case with partially observed $\mathbf{s}$ which is more applicable in real-world settings nor do they address the problem of inactive stochastic units inherent in deep latent variable models.

Furthermore, even though the formulation of the graphical model encourages separation between the sensitive variable and the latent representation, some sensitive information can remain if this information is correlated with the prediction task. Therefore, an additional regularization term is necessary to further enforce fairness of the representation. We follow this line of thinking and explore the mutual information as a fairness regularizer to ensure that the sensitive information is removed completely.

The contribution of the paper is twofold:
\begin{itemize}
    \item[--] We propose a new deep generative model for learning fair representations and empirically show that it outperforms the VFAE when $\mathbf{s}$ is observed partially and performs on par when the sensitive variable is fully observed.
    \item[--] We introduce the mutual information as a new fairness regularizer that is better suited for the realistic case where the sensitive variables are partially observed.
\end{itemize}

\section{Fair Deep Generative Models}

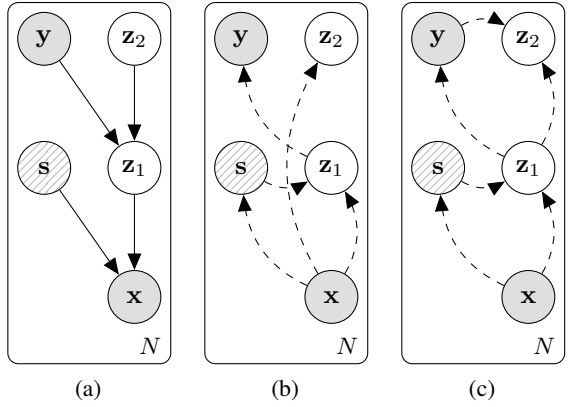
\begin{figure}%
\centering
\subfigure[]{%
\label{fig:first_gen}%
\begin{tikzpicture}

  \node[obs]                               (x) {$\mathbf{x}$};
  \node[semi, above=of x, xshift=-1.2cm] (s) {$\mathbf{s}$};
  \node[latent, above=of x]  (z1) {$\mathbf{z}_1$};
  \node[obs, above=of s]  (y) {$\mathbf{y}$};
  \node[latent, above=of z1] (z2) {$\mathbf{z}_2$};

  \edge {s, z1} {x} ;
  \edge {y, z2} {z1} ;

  \plate {} {(x)(y)(z1)(z2)(s)} {$N$} ;

\end{tikzpicture}

}%
\quad
\subfigure[]{%
\label{fig:second_var}%
\begin{tikzpicture}

  \node[obs]                               (x) {$\mathbf{x}$};
  \node[semi, above=of x, xshift=-1.2cm] (s) {$\mathbf{s}$};
  \node[latent, above=of x]  (z1) {$\mathbf{z}_1$};
  \node[obs, above=of s]  (y) {$\mathbf{y}$};
  \node[latent, above=of z1] (z2) {$\mathbf{z}_2$};

  \edge [bend right, dashed] {s, x} {z1} ;
  \edge [bend left, dashed] {x} {z2} ;
  \edge [bend left, dashed] {z1} {y} ;
  \edge [bend left, dashed] {x} {s} ;

  \plate {} {(x)(y)(z1)(z2)(s)} {$N$} ;

\end{tikzpicture}

}%
\quad
\subfigure[]{%
\label{fig:third_var}%
\begin{tikzpicture}

  \node[obs]                               (x) {$\mathbf{x}$};
  \node[semi, above=of x, xshift=-1.2cm] (s) {$\mathbf{s}$};
  \node[latent, above=of x]  (z1) {$\mathbf{z}_1$};
  \node[obs, above=of s]  (y) {$\mathbf{y}$};
  \node[latent, above=of z1] (z2) {$\mathbf{z}_2$};

  \edge [bend right, dashed] {s, x} {z1} ;
  \edge [bend left, dashed] {y} {z2} ;
  \edge [bend right, dashed] {z1} {z2} ;
  \edge [bend left, dashed] {z1} {y} ;
  \edge [bend left, dashed] {x} {s} ;

  \plate {} {(x)(y)(z1)(z2)(s)} {$N$} ;

\end{tikzpicture}

}%
\caption{(a) Generative part of both models. (b) Variational part H-VFAE. (c) Variational part VFAE.}
\label{fig:graphicalmodel}
\end{figure}

Learning fair representations aims at becoming invariant to nuisance or sensitive factors while retaining as much of the remaining relevant information as possible \cite{louizos2015variational, zemel2013learning}. From the probabilistic modeling point of view, the problem could be formulated in terms of a set of independent factors working on the input $\mathbf{x} \in \mathcal{X}$, $\mathcal{X}$ is a $D$-dimensional discrete or continuous space, namely, the (partially)-observed discrete sensitive (nuisance) variable $\mathbf{s} \in \mathcal{S}$, typically $\mathcal{S} = \{0,1\}$, and the continuous unobserved latent variable $\mathbf{z}_1 \in \mathbb{R}^{M_1}$. Additionally, since the goal is to learn features that are invariant to $\mathbf{s}$ without losing information about the label $\mathbf{y}$, a hierarchy of latent variables could be introduced. In this paper, we assume a second layer of latent variables $\mathbf{z}_2 \in \mathbb{R}^{M_2}$. All the label independent noise inherent in $\mathbf{x}$ is modeled in the hierarchical latent representation, while also allowing the model to correlate the discrete label information $\mathbf{y}$ with the invariant features $\mathbf{z}_1$. As a result, the following generative process could be considered:
\begin{align}
\mathbf{y} &\sim \text{Cat}(\mathbf{y}) \\ 
\mathbf{z}_2 &\sim p(\mathbf{z}_2) \\
\qquad \mathbf{z}_1 &\sim p_\theta(\mathbf{z}_1|\mathbf{z}_2,\mathbf{y}) \\
\mathbf{x} &\sim p_\theta(\mathbf{x}|\mathbf{z}_1,\mathbf{s}) ,
\end{align}
where $\text{Cat}(\cdot)$ denotes the categorical distribution, see Figure \ref{fig:graphicalmodel} for the probabilistic graphical model. This formulation can be recast as an inference problem where the objective is to learn the posterior $p(\mathbf{z}_1, \mathbf{z}_2, \mathbf{y}|\mathbf{x}, \mathbf{s})$ in the case of observed $\mathbf{s}$, and $p(\mathbf{z}_1, \mathbf{z}_2, \mathbf{y}, \mathbf{s}|\mathbf{x})$ in the case of unobserved $\mathbf{s}$. Since calculating the true posterior is infeasible, we will use variational inference and the methodology of variational auto-encoders (VAE) \cite{kingma2013auto}.

\section{Variational Fair Auto-Encoder}
Depending on assumed dependencies among random variables in the variational posterior, the application of variational inference to the generative process presented in the previous section may result in different VAE architectures. \citet{louizos2015variational} assumed $\mathbf{s}$ is always given and, thus, they proposed to factorize the variational posterior $q_\phi(\mathbf{z}_1, \mathbf{z}_2, \mathbf{y}|\mathbf{x}, \mathbf{s})$ as $q_\phi(\mathbf{z}_1|\mathbf{x},\mathbf{s})q_\phi(\mathbf{y}|\mathbf{z}_1) q_\phi(\mathbf{z}_2|\mathbf{z}_1,\mathbf{y})$, see Figure \ref{fig:third_var}. The final model is defined as follows:
\begin{align}
q_\phi(\mathbf{z}_1|\mathbf{x},\mathbf{s}) &= \mathcal{N}(\mathbf{z}_1|\boldsymbol{\mu}_\phi(\mathbf{x},\mathbf{s}), \boldsymbol{\sigma}_\phi(\mathbf{x},\mathbf{s}))\\
q_\phi(\mathbf{y}|\mathbf{z}_1) & = \text{Cat}(\mathbf{y}|\boldsymbol{\pi}_\phi(\mathbf{z}_1))\\
q_\phi(\mathbf{z}_2|\mathbf{z}_1,\mathbf{y}) &= \mathcal{N}(\mathbf{z}_2|\boldsymbol{\mu}_\phi(\mathbf{z}_1,\mathbf{y}), \boldsymbol{\sigma}_\phi(\mathbf{z}_1,\mathbf{y})) \\
p(\mathbf{z}_2) &= \mathcal{N}(\mathbf{z}_{2}|\mathbf{0}, \mathbf{I})\\
p_\theta(\mathbf{z}_1|\mathbf{z}_2,\mathbf{y}) &= \mathcal{N}(\mathbf{z}_1|\boldsymbol{\mu}_\phi(\mathbf{z}_2,\mathbf{y}), \boldsymbol{\sigma}_\phi(\mathbf{z}_2,\mathbf{y}))\\
p_\theta(\mathbf{x}|\mathbf{z}_1,\mathbf{s}) &= f_\theta(\mathbf{z}_1,\mathbf{s}) ,
\end{align}
where all distributions are parameterized by neural networks, and $f_\theta(\mathbf{z}_1,\mathbf{s})$ is a distribution suited for the data that is modeled.  Following this formulation, we aim at maximizing the variational (evidence) lower bound on $\ln p(\mathbf{x}|\mathbf{s})$ (ELBO):
\begin{align}
\mathcal{L}&_s(\mathbf{x}, y, \mathbf{s}) =\mathbb{E}_{q_\phi(\mathbf{z}_1|\mathbf{x},\mathbf{s})}[\ln p_\theta(\mathbf{x}|\mathbf{z}_1,\mathbf{s}) -\\
&\text{KL}\big(q_\phi(\mathbf{z}_2|\mathbf{z}_1, \mathbf{y})\,  || \, p(\mathbf{z}_2)\big)] +\\
&\mathbb{E}_{q_\phi(\mathbf{z}_1, \mathbf{z}_2|\mathbf{x},\mathbf{s},\mathbf{y})}[\ln p_\theta(\mathbf{z}_1|\mathbf{z}_2,\mathbf{y}) - \ln q_\phi(\mathbf{z}_1|\mathbf{x},\mathbf{s}) ] +\\
&\alpha \mathbb{E}_{q_\phi(\mathbf{z}_1|\mathbf{x},\mathbf{s})}[\ln q_\phi(\mathbf{y}|\mathbf{z}_1)] 
\end{align}
where $\alpha>0$ is an additional parameter to control the influence of the classifier during training.
The ELBO can be jointly optimized with respect to the parameters $\phi, \theta$ of the inference and generative model, respectively, using the reparameterization trick \cite{kingma2013auto}. We refer to this model as the Variational Fair Auto-Encoder (VFAE).


\section{Hierarchical VampPrior VFAE}

The VFAE is shown to be successful in learning fair representations
\cite{louizos2015variational}. However, it has been shown that in general deep VAEs suffer from the inactive latent variable problem \cite{sonderby2016ladder}, following from a top-down multi-layered generative process while the variational part is bottom-up. Therefore, we propose to change the variational part of the VFAE with inputs fed directly to the deepest layer such that the final encoder changes to $q_\phi(\mathbf{z}_2|\mathbf{x})$. This has the effect of enforcing a dependency between the data and the latent units at the deepest level during the generative process, preventing the latent units from regularization towards the prior, i.e. setting $q_\phi(\mathbf{z}|\mathbf{x}) = p(\mathbf{z)}$.

Furthermore, this formulation allows for easy integration of a recently proposed powerful prior, the Variational Mixture of Posteriors Prior (VampPrior) with a hierarchical architecture \cite{VampPrior}.  We hypothesize that the quality of the VFAE could be improved by utilizing a different family of variational posteriors coupled with a powerful prior over the latent representation utilizing the new hierarchical structure.

Eventually, we end up with a different structure of the variational posterior (see Figure \ref{fig:second_var} for the probabilistic graphical model):
\begin{equation}
    q_\phi(\mathbf{z}_1, \mathbf{z}_2, \mathbf{y}|\mathbf{x}, \mathbf{s}) = q_\phi(\mathbf{z}_1|\mathbf{x},\mathbf{s}) q_\phi(\mathbf{y}|\mathbf{z}_1) q_\phi(\mathbf{z}_2|\mathbf{x}) .
\end{equation}
Now, we can consider the problem of finding the prior that optimizes the lower bound given the data. The solution is simply the aggregated posterior \cite{hoffman2016elbo}:
\begin{equation}
    p^*_{\lambda}(\mathbf{z}) = \frac{1}{N}\sum^N_{n=1}q_\phi(\mathbf{z}|\mathbf{x}_n) .
\end{equation}
This could, however, lead to overfitting and would be very expensive to compute for every training iteration. A computationally efficient alternative, which also prevents from overfitting by restricting $K \ll N$, is an approximation using a mixture of variational posteriors with learnable pseudo-inputs \cite{VampPrior}:
\begin{equation}
    p_\lambda(\mathbf{z}) = \frac{1}{K}\sum^K_{k=1} q_\phi(\mathbf{z}|\mathbf{u}_k) ,
\end{equation}
where $K$ is the number of pseudo-inputs and $\mathbf{u}_k$ denotes the $k$-th pseudo-input that is of the same dimension as the input. 

The final model is defined as follows:
\begin{align}
q_\phi(\mathbf{z}_1|\mathbf{x},\mathbf{s}) &= \mathcal{N}(\mathbf{z}_1|\boldsymbol{\mu}_\phi(\mathbf{x},\mathbf{s}), \boldsymbol{\sigma}_\phi(\mathbf{x},\mathbf{s}))\\
q_\phi(\mathbf{y}|\mathbf{z}_1) & = \text{Cat}(\mathbf{y}|\boldsymbol{\pi}_\phi(\mathbf{z}_1))\\
q_\phi(\mathbf{z}_2|\mathbf{x}) &= \mathcal{N}(\mathbf{z}_2|\boldsymbol{\mu}_\phi(\mathbf{x}), \boldsymbol{\sigma}_\phi(\mathbf{x})) \\
p_\lambda(\mathbf{z}_2) &= \frac{1}{K}\sum^K_{k=1} q_\phi(\mathbf{z}_2|\mathbf{u}_{k}) \\
p_\theta(\mathbf{z}_1|\mathbf{z}_2,\mathbf{y}) &= \mathcal{N}(\mathbf{z}_1|\boldsymbol{\mu}_\phi(\mathbf{z}_2,\mathbf{y}), \boldsymbol{\sigma}_\phi(\mathbf{z}_2,\mathbf{y}))\\
p_\theta(\mathbf{x}|\mathbf{z}_1,\mathbf{s}) &= f_\theta(\mathbf{z}_1,\mathbf{s})
\end{align}
The objective function is the ELBO in the following form:
\begin{align}
\mathcal{L}_s(\mathbf{x},& y, \mathbf{s}) = \mathbb{E}_{q_\phi(\mathbf{z}_1|\mathbf{x},\mathbf{s})}[\ln p_\theta(\mathbf{x}|\mathbf{z}_1,\mathbf{s})] -\\
& \text{KL}\big(q_\phi(\mathbf{z}_2|\mathbf{x})\,  || \, p_\lambda(\mathbf{z}_2)\big) -\\
& \mathbb{E}_{q_\phi(\mathbf{z}_2|\mathbf{x})}[ \text{KL}(q_\phi(\mathbf{z}_1|\mathbf{x}, \mathbf{s}) \, || \, p_\theta(\mathbf{z}_1|\mathbf{z}_2,\mathbf{y}) )] + \\ 
& \alpha \mathbb{E}_{q_\phi(\mathbf{z}_1|\mathbf{x},\mathbf{s})}[\ln q_\phi(\mathbf{y}|\mathbf{z}_1)] .
\end{align}
We refer to this model as Hierarchical VampPrior Variational Fair Auto-Encoder (H-VFAE + VP).

Alternatively, we can consider a simpler case where the VampPrior is replaced by the standard Gaussian prior, $p_{\lambda}(\mathbf{z}_{2}) = \mathcal{N}(\mathbf{z}_{2}|\mathbf{0}, \mathbf{I})$. We will call this model the Hierarchical Variational Fair Auto-Encoder (H-VFAE).

\section{Encouraging learning fair representations}
If the sensitive variable is correlated with the prediction task, information about $\mathbf{s}$ can still remain in the latent representation $\mathbf{z}_1$. To remove this information, two fairness penalties are discussed, which can easily be added to the lower bound as a regularizer.

\subsection{MMD regularizer}
\citet{louizos2015variational} originally proposed to use the Maximum Mean Discrepancy (MMD) measure to regularize the marginal $q_\phi(\mathbf{z}_1|\mathbf{s})$. The rationale behind applying the MMD is that it compares statistics of two samples, and if they are similar, the MMD indicates that they were drawn from the same distribution. The distance between the empirical statistics $\varphi$ of two datasets can be computed in the following manner:
\begin{equation}
    \| \frac{1}{N_0}\sum^{N_0}_{i=1}\varphi(\mathbf{z}_0) - \frac{1}{N_1}\sum^{N_1}_{i=1}\varphi(\mathbf{z}_1)\|^2 .
\end{equation}
An unbiased MMD estimator \cite{gretton2007kernel} is obtained by expanding the square and is solely composed of inner products on which the kernel trick can be applied:
\begin{align}
\ell&_{MMD} = \mathbb{E}_{p(\mathbf{z}_0), p(\mathbf{z}'_0)}[k(\mathbf{z}_0,\mathbf{z}'_0)] + \\
&\mathbb{E}_{q(\mathbf{z}_1), q(\mathbf{z}'_1)}[k(\mathbf{z}_1,\mathbf{z}'_1)] - 2\mathbb{E}_{p(\mathbf{z}_0), q(\mathbf{z}_1)}[k(\mathbf{z}_0,\mathbf{z}_1)] .
\end{align}
Optimizing for the MMD regularizer has the effect of matching the moments of marginal distributions $q_\phi(\mathbf{z}_1|\mathbf{s}= 0)$ and $q_\phi(\mathbf{z}_1|\mathbf{s}= 1)$, while still allowing individual elements to differ. In our case, the MMD regularizer is the following:
\begin{align}
\ell_{MMD} = \mathbb{E}_{ \tilde{p}(\mathbf{x})} \Big{[} \| &\mathbb{E}_{q_\phi(\mathbf{z}_1|\mathbf{x},\mathbf{s}=0)}[\varphi(\mathbf{z}_1)] - \\ &\mathbb{E}_{q_\phi(\mathbf{z}_1|\mathbf{x},\mathbf{s}=1)}[\varphi(\mathbf{z}_1)]\| ^2 \Big{]},
\end{align}
where $\tilde{p}(\mathbf{x})$ denotes the empirical distribution.

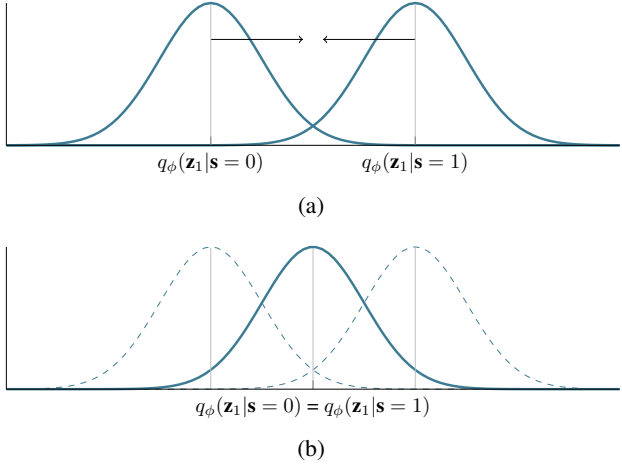
\begin{figure}%
\centering
\subfigure[]{%
\label{fig:first_MMD}%
\resizebox{1.0\linewidth}{!}{\begin{tikzpicture}
\begin{axis}[
  no markers, domain=\domainstart:\domainend, samples=100,
  axis lines*=left,
  xticklabels={$q_\phi(\textbf{z}_1|\textbf{s}=0)$, $q_\phi(\textbf{z}_1|\textbf{s}=1)$},
  height=\height, width=\width,
  xtick={\meanone, \meantwo}, ytick=\empty,
  enlargelimits=false, clip=false, axis on top,
  ]
  \addplot [very thick,cyan!50!black] {gauss(\meanone,\sigmaone)};
  \addplot [very thick,cyan!50!black] {gauss(\meantwo,\sigmatwo)};

\draw[yshift=\yshift, ->] (\meanone * 100,0) -- (\mmdleft * 100,0);
\draw[yshift=\yshift, ->] (\meantwo * 100,0) -- (\mmdright * 100,0);
\draw[color=lightgray](\meanone * 100,0) -- (\meanone * 100, \meanheight);
\draw[color=lightgray](\meantwo * 100,0) -- (\meantwo * 100, \meanheight);
\end{axis}

\end{tikzpicture}}}
\quad
\subfigure[]{%
\label{fig:second_MMD}%
\resizebox{1.0\linewidth}{!}{\begin{tikzpicture}
\begin{axis}[
  no markers, domain=\domainstart:\domainend, samples=100,
  axis lines*=left,
  xticklabels={$q_\phi(\textbf{z}_1|\textbf{s}=0)$ = $q_\phi(\textbf{z}_1|\textbf{s}=1)$},
  height=\height, width=\width,
  xtick={\meanthree}, ytick=\empty,
  enlargelimits=false, clip=false, axis on top,
  ]
  \addplot [dashed,cyan!50!black] {gauss(\meanone,\sigmaone)};
  \addplot [dashed,cyan!50!black] {gauss(\meantwo,\sigmatwo)};
  \addplot [very thick,cyan!50!black] {gauss(\meanthree, \sigmathree)};

\draw[color=lightgray](\meanone * 100,0) -- (\meanone * 100, \meanheight);
\draw[color=lightgray](\meantwo * 100,0) -- (\meantwo * 100, \meanheight);
\draw[color=lightgray](\meanthree * 100,0) -- (\meanthree * 100, \meanheight);
\end{axis}

\end{tikzpicture}}}
\vskip -3mm
\caption{A schematic presentation of the behavior of the MMD regularizer.}
\label{fig:MMD}%
\end{figure}

The behavior of the MMD regularizer is schematically presented in Figure \ref{fig:MMD}. Two marginal distributions are matched by matching their respective moments. 

\subsubsection{Fast MMD regularizer}
To prevent computing the expensive full MMD estimator, random kitchen sinks \cite{rahimi2009weighted} can be used to compute the feature expansion $\varphi(\mathbf{z})$ to serve as an approximation to the MMD regularizer. The idea is to draw a random matrix $\mathbf{W} \in \mathbb{R}^{M \times K}$, with $M$ as the dimensionality of $\mathbf{z}$ and $K$ as the number of random features, where each entry is drawn from a standard isotropic Gaussian. Additionally, a $M$-dimensional uniform random vector $\mathbf{b}$ is drawn with entries in $[0, 2\pi]$. The feature expansion can be then computed as follows \cite{louizos2015variational}:
\begin{equation}
\varphi_\mathbf{W}(\mathbf{z}) =\sqrt{\frac{2}{D}\cos}\Big( \sqrt{\frac{2}{\gamma}}\mathbf{z} \mathbf{W} + \mathbf{b}\Big) ,    
\end{equation}
where $\gamma = 2M$.
The inner product of these randomized feature expansions converges to a kernel function given an increasing number of features.

\subsection{Mutual information regularizer}
\begin{figure}%
\centering
\subfigure[]{%
\label{fig:first_MI}%
\resizebox{1.0\linewidth}{!}{\begin{tikzpicture}
\begin{axis}[
  no markers, domain=\domainstart:\domainend, samples=100,
  axis lines*=left,
  xticklabels={$q_\phi(\textbf{z}_1|\textbf{x}{,\,}\textbf{s}=0)$, $q_\phi(\textbf{z}_1|\textbf{x}{,\,}\textbf{s}=1)$},
  height=\height, width=\width,
  xtick={\meanone, \meantwo}, ytick=\empty,
  enlargelimits=false, clip=false, axis on top,
  ]
\addplot [very thick,cyan!50!black] {gauss(\meanone,\sigmaone)};
\addplot [very thick,cyan!50!black] {gauss(\meantwo,\sigmatwo)};
\addplot [densely dotted,red!50!black, thick] {mmgauss(\meanone,\sigmaone, \meantwo,\sigmatwo, \mixone)};
\draw[yshift=\yshift, ->] (\meantwo * 100,0) -- (\mileft * 100,0);
\draw[color=lightgray](\meanone * 100,0) -- (\meanone * 100, \meanheight);
\draw[color=lightgray](\meantwo * 100,0) -- (\meantwo * 100, \meanheight);
\draw[color=lightgray](\meanmixtureone * 100,0) -- (\meanmixtureone * 100, \mixheight);

\end{axis}
\end{tikzpicture}}}
\quad
\subfigure[]{%
\label{fig:second_MI}%
\resizebox{1.0\linewidth}{!}{\begin{tikzpicture}
\begin{axis}[
  no markers, domain=\domainstart:\domainend, samples=100,
  axis lines*=left,
  xticklabels={$q_\phi(\textbf{z}_1|\textbf{x}{,\,}\textbf{s}=0)$, $q_\phi(\textbf{z}_1|\textbf{x}{,\,}\textbf{s}=1)$},
  height=\height, width=\width,
  xtick={\meanone, 3.3}, ytick=\empty,
  enlargelimits=false, clip=false, axis on top,
  ]
\addplot [very thick,cyan!50!black] {gauss(\meanone,\sigmaone)};
\addplot [very thick,cyan!50!black] {gauss(\meanthree,\sigmatwo)};
\addplot [densely dotted,red!50!black, thick] {mmgauss(\meanone,\sigmaone, \meanthree,\sigmatwo, \mixone)};
\draw[color=lightgray](\meanone * 100,0) -- (\meanone * 100, \meanheight);
\draw[color=lightgray](\meanthree * 100,0) -- (\meanthree * 100, \meanheight);
\draw[color=lightgray](\meanmixturethree * 100,0) -- (\meanmixturethree * 100, \mixheighttwo);

\end{axis}

\end{tikzpicture}}}
\quad
\subfigure[]{%
\label{fig:third_MI}%
\resizebox{1.0\linewidth}{!}{\begin{tikzpicture}
\begin{axis}[
  no markers, domain=\domainstart:\domainend, samples=100,
  axis lines*=left,
  xticklabels={$q_\phi(\textbf{z}_1|\textbf{x}{,\,}\textbf{s}) = q_\phi(\textbf{z}_1|\textbf{x})$},
  height=\height, width=\width,
  xtick={\meanthree}, ytick=\empty,
  enlargelimits=false, clip=false, axis on top,
  grid = major
  ]
\addplot [very thick,cyan!50!black] {gauss(\meanthree,\sigmaone)};
\addplot [densely dotted,red!50!black, thick] {gauss(\meanthree,\sigmaone)};

\end{axis}

\end{tikzpicture}}}
\vskip -3mm
\caption{A schematic representation of the behavior of the MI regularizer.}
\label{fig:MI}
\end{figure}
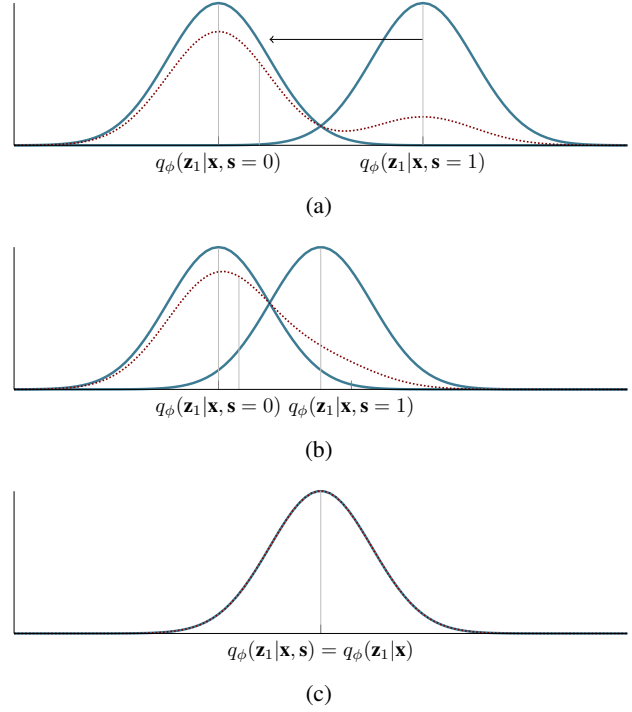

Another manner of enhancing fairness is to force independence between the representation and the sensitive variable. A natural candidate for this purpose is the mutual information, which represents a measure of mutual dependence between two random variables. In our case we are interested in the conditional mutual information between $\mathbf{z}_{1}$ and $\mathbf{s}$ for given $\mathbf{x}$, that is:
\begin{equation}
MI(\mathbf{z}_1, \mathbf{s}|\mathbf{x}) =\mathbb{E}_{\tilde{p}(\mathbf{x})q_\phi(\mathbf{z}_1, \mathbf{s}|\mathbf{x})}\Big[\ln \frac{q_\phi(\mathbf{z}_1, \mathbf{s}|\mathbf{x})}{q_\phi(\mathbf{z}_1|\mathbf{x})q_\phi(\mathbf{s}|\mathbf{x})}\Big] .
\end{equation}
Furthermore assuming that the posterior factorizes as $q_\phi(\mathbf{s}|\mathbf{x})q_\phi(\mathbf{z}_1|\mathbf{x}, \mathbf{s})$ we obtain an easily computable and differentiable estimator:
\begin{align}
\ell&_{MI}=\mathbb{E}_{\tilde{p}(\mathbf{x})q_\phi(\mathbf{z}_1, \mathbf{s}|\mathbf{x})}\Big[\ln \frac{q_\phi(\mathbf{z}_1|\mathbf{x},\mathbf{s})}{q_\phi(\mathbf{z}_1|\mathbf{x})}\Big]   \\
&=  \mathbb{E}_{\tilde{p}(\mathbf{x})q_\phi(\mathbf{z}_1, \mathbf{s}|\mathbf{x})}\Big[\ln \frac{q_\phi(\mathbf{z}_1|\mathbf{x},\mathbf{s})}{\sum_s q_\phi(\mathbf{s}|\mathbf{x}) q_\phi(\mathbf{z}_1|\mathbf{x}, \mathbf{s})}\Big] ,
\end{align}
which can be approximated using Monte Carlo samples. Typically, $\mathbf{s}$ is low-dimensional (e.g., it is binary), hence, calculating the mixture distribution in the denominator is easily tractable. Stochastic gradient ascent can now be performed on $\nabla_{\phi} \ell_{MI}$ to regularize the encoder $q_\phi(\mathbf{z}_1|\mathbf{x},\mathbf{s})$. The behavior of the Mutual Information regularizer is schematically presented in Figure \ref{fig:MI}. Notice that in contrast to the MMD regularizer, here we try to match each mode separately to the mixture and the optimal solution is attained when both modes overlap. In other words, the encoder does not use the information about the sensitive variable solely if it produces the same distribution for any value of $\mathbf{s}$.

\section{Partial supervision of the sensitive variable}

Typically, it is assumed that the sensitive variable is fully observable. However, in many real-life applications $\mathbf{s}$ is only partially-observable. In this case, the generative approach allows the model to infer the sensitive variable so that all data could be used during training.

The supervised model can be easily extended to cope with these examples where no sensitive variables are provided by adding a variational categorical distribution $q_\phi(\mathbf{s}|\mathbf{x})$ to the model. Besides $\mathcal{L}_s$ we now jointly optimize the unsupervised lower bound where we can sum over all possible values of $\mathbf{s}$ or we can use differentiable samples from the random discrete node $q_\phi(\mathbf{s}|\mathbf{x})$ obtained by the reparameterization trick using the concrete distribution \cite{jang2016categorical, maddison2016concrete}:
\begin{equation}
    \mathcal{L}_u(\mathbf{x}, y) = \mathbb{E}_{q_\phi(\mathbf{s}|\mathbf{x})}\big[ \mathcal{L}_s(\mathbf{x},y, \mathbf{s})\big] - KL\big( q_\phi(\mathbf{s}|\mathbf{x})\, ||\, p(\mathbf{s})\big) ,
\end{equation}
where the following distributions are introduced:
\begin{align}
    q_\phi(\mathbf{s}|\mathbf{x}) &= \text{Cat}(\mathbf{s}|\boldsymbol{\pi}_\phi(\mathbf{x})) \\
    p(\mathbf{s}) &= \text{Cat}(\mathbf{s}|\pi) ,
\end{align}
and $\pi$ denote a priori probabilities of sensitive variables.

Eventually, we can combine both the supervised and unsupervised objectives together that gives our final objective function for given training data $\mathcal{D}$:
\begin{align}
\mathcal{L}(\mathcal{D}) =& \mathbb{E}_{\tilde{p}(\mathbf{x},y,\mathbf{s})}[ \mathcal{L}_s(\mathbf{x}, y, \mathbf{s}) ] + \mathbb{E}_{\tilde{p}(\mathbf{x,s})}[ \mathcal{L}_u(\mathbf{x}, y) ] + \\
& \mathbb{E}_{\tilde{p}(\mathbf{x,s})}[-\ln q_\phi(\mathbf{s}|\mathbf{x})] - \mathbb{E}_{\tilde{p}(\mathbf{x},y)}[ \lambda \ell(\mathbf{z}_1) ],
\end{align}
where $\tilde{p}(\cdot)$ denotes the empirical distribution, $\lambda > 0$, and $\ell(\mathbf{z}_1)$ is either the MMD regularizer or the MI regularizer.

\section{Experiments}
\subsection{Datasets}
Experiments were run on the German and Adult datasets with the same training, validation and test splits as used in
\cite{zemel2013learning}. The German dataset consists of credit data and the objective is to predict if a person has a good or bad credit rating. The sensitive variable here is the age of the individual. The Adult income dataset consists of census data and the prediction task is to determine whether a person makes over 50.000 dollars a year. The sensitive variable here is the gender.
We binarized both datasets and used a Bernoulli distribution for the final decoder, i.e. $p_\theta(\mathbf{x}|\mathbf{z}_1, \mathbf{s}) = \text{Bern}(\mathbf{x}|\boldsymbol{\pi}_\phi(\mathbf{z}_1, \mathbf{s}))$ similarly to \cite{louizos2015variational}. 

\subsection{Settings}
The same neural network architectures as in \cite{louizos2015variational} were used for all experiments. For the small German dataset, a hidden layer of 60 units was used for all encoders and decoders, with a stochastic latent dimensionality of 30 units. For the Adult income dataset, 100 hidden units were used for all encoders and decoders, while the dimensionality of the latent space was increased to 50. 

Like \cite{louizos2015variational} we took $\alpha = 1$ for the supervised setting, while $\beta$ was cross-validated due to the varying nature of the strength of the regularizers. For the partially-supervised setting, we set $\alpha = 20$ as it was observed that the partially-supervised regularized models were well suited to handle the increased dependency on the classification error. Optimization was done with the Adam optimizer \cite{kingma2014adam}, where the default settings were used.

\subsection{Evaluation}
The primary goal of the paper is fair classification, therefore, all models need to be evaluated with respect to the classification accuracy and with respect to information being available about the sensitive variables.

To measure the information about the sensitive variables remaining in the predictive features, a logistic regression classifier was trained to predict the state of our sensitive variable given the features from the variational posterior $q_\phi(\mathbf{z}_1|\mathbf{x},\mathbf{s})$.

Furthermore, since another objective of fair classification is group fairness, ensuring equal treatment between different groups, the probabilistic discriminative metric from \cite{louizos2015variational} is used: 
\begin{equation}
\text{DS} = \Big|\frac{\sum_{n|s_n=0}p(\hat{y}_n)}{N_{s=0}} -\frac{\sum_{n|s_n=1}p(\hat{y}_n)}{N_{s=1}}\Big| ,
\end{equation}
This metric has the simple interpretation of measuring the difference in classification predictions between different groups.

Both models were tested on the full supervision and partial supervision of $\mathbf{s}$, where the fraction of observed sensitive variables was set to 0.05. To test the regularization penalties we evaluate the models with $\ell_{MI}, \,\ell_{MMD}$ and no regularizer.

Additionally, the influence of the VampPrior was isolated by also providing a baseline of our proposed model with the standard Gaussian prior instead of a richer one, denoted by H-VFAE in our experiments.

{\renewcommand{\arraystretch}{1.15}
\begin{table*}[!htb]
\caption{Results on the fully supervised case. Best results of methods with fairness regularization in bold.}
\label{supervised}
\vskip 0.15in
\begin{center}
\begin{small}
\begin{sc}
\begin{tabular}{llcccrr}
\hline
\abovespace\belowspace
Model & German Y & Adult Y & German S & Adult S & German DS & Adult DS \\
\hline
\abovespace
Random & 71.1 & 75.0 & 80.1 & 67.0 & - & -  \\
VFAE & 72.4 & 82.0 & 80.1 & 66.1 & 2.7 & 5.4   \\
H-VFAE & 72.5 & 80.5 & 80.1 & 68.3 & 10.1 & 11.3 \\
H-VFAE + VP & 72.4 & 81.9 & 80.1 & 67.2 & 5.5 & 7.9 \\
\hline
VFAE + MMD & 72.7 & 81.3 & 80.1 & 67.4 & \textbf{0.6} & \textbf{2.5}   \\
H-VFAE + MMD & 74.2 & 81.0 & 80.1 & 67.2 & 7.2 & 8.6 \\
H-VFAE + VP + MMD & \textbf{74.4} & \textbf{82.2} & 80.1 & 67.2 & 1.6 & 3.3 \\
\hline
VFAE + MI & 72.9 & 81.6 & 80.1 & 67.4 & 4.2 & 3.4   \\
H-VFAE + MI & 73.6 & 82.1 & 80.1 & 67.4 & 5.1 & 5.6 \\
H-VFAE + VP + MI & 73.4 & 82.1 & 80.1 & 67.3 & 3.7 & \textbf{2.5} \\
\hline
\end{tabular}
\end{sc}
\end{small}
\end{center}
\vskip -0.1in
\end{table*}}

{\renewcommand{\arraystretch}{1.15}
\begin{table*}[!htb]
\caption{Results on the partially supervised case. Best results of methods with fairness regularization in bold.}
\label{semisupervised}
\vskip 0.15in
\begin{center}
\begin{small}
\begin{sc}
\begin{tabular}{llcccrr}
\hline
\abovespace\belowspace
Model & German Y & Adult Y & German S & Adult S & German DS & Adult DS \\
\hline
\abovespace
Random & 71.1 & 75.0 & 80.1 & 67.0 & - & -  \\
VFAE & 75.5 & 84.8 & 80.1 & 69.7 & 8.6 & 11.4     \\
H-VFAE + VP & 75.1 & 84.5 & 80.1 & 69.4 & 8.8 & 10.7    \\
\hline
VFAE + MMD & 73.4 & 81.5 & 80.1 & 67.4 & 3.4 & 8.1    \\
H-VFAE + VP + MMD & 73.4 & 81.7 & 80.1 & 67.4 & 3.2 & 6.1   \\
\hline
VFAE + MI & 72.8 & 82.0 & 80.1 & 67.4 & 3.3 & 5.6 \\
H-VFAE + VP + MI & \textbf{74.1} & \textbf{82.3} & 80.1 & 67.4 & \textbf{3.1} & \textbf{4.9} \\
\hline
\end{tabular}
\end{sc}
\end{small}
\end{center}
\vskip -0.1in
\end{table*}}

\subsection{Experiment with the full supervision of $\mathbf{s}$}

In the first experiment, our model and the newly proposed regularizer are evaluated on the case with fully observed sensitive variables.\footnote{In order to have comparable results to the original paper on VFAE, we used the same experiment setting as in \cite{louizos2015variational}.} The results are presented in Table \ref{supervised}.

First of all, we notice that without any regularization the proposed family of variational posteriors performs similarly to the VFAE in terms of the classification accuracy on $y$, however, it performs worst on the DS metric. It is also worth to note that our model greatly benefits from the VampPrior. 


The benefits of the VampPrior on our new architecture are easily observed by noting that the prediction accuracy on average increases on both datasets while still having a regularized effect, resulting in a lower amount of information available about $\mathbf{s}$.

\begin{figure}[H]
    \centering
    \includegraphics[width=1\columnwidth]{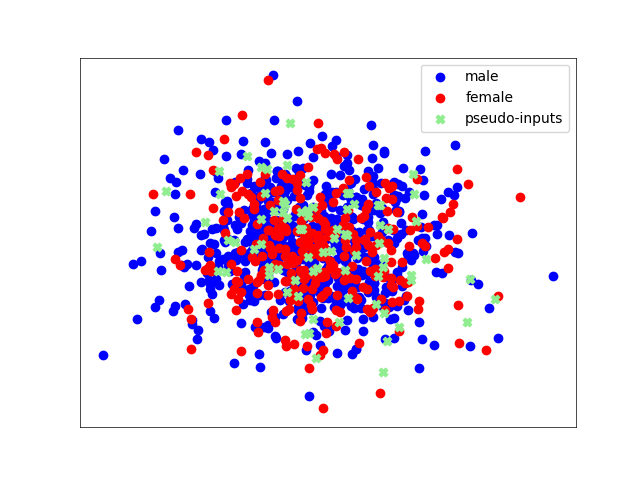}
    \vskip -7mm
    \caption{The 2D latent space visualization for $\mathbf{z}_1$ in the Adult dataset. The colors correspond to gender values and the crosses represent means for components of the VampPrior.}
    \label{fig:pseudo}
\end{figure}

Moreover, the effect of applying the VampPrior is presented in Figure \ref{fig:pseudo} (the crosses represent means of the components $q_{\phi}(\mathbf{z}_{1}|\mathbf{u}_{k})$). Notice that the prior is highly multi-modal and it covers the latent space in places where the encoder places $\mathbf{z}_1$ for given $\mathbf{x}$ and $\mathbf{s}$. If the standard Gaussian prior is used, the encoder would be forced to put most of the points close to the origin. Also note that the latent representations are almost indistinguishable for the two groups.

Additionally, the new model with the VampPrior outperforms the original VFAE on predictive capabilities on both datasets while keeping the information about $\mathbf{s}$ similar. Furthermore, all models are as good as invariant against classification with respect to $\mathbf{s}$ on the features $\mathbf{z}_1$.

If we look at the regularization penalties, it is clear that both the MMD and MI penalties have a significant regularizing effect on the information retained in $\mathbf{s}$. The MMD regularizer is shown to be a marginally better fit for the supervised case with slightly lower scores on $\mathbf{s}$ while retaining the same classification accuracy. A possible explanation for that is that for sufficiently large training samples, the MMD is better approximated than the MI regularizer. Another explanation is that we might need to perform more thorough hyperparameter searches. Even after our quite exhaustive search, it might be the case that there is a warm spot for which the MI regularizer might give better results in terms of the DS metric. 

All in all, the H-VFAE+VP with the MMD regularizer seems to give the best trade-off between predictive accuracy and losing information about $\mathbf{s}$ in the supervised case, outperforming all the other models.


\subsection{Experiment with the partial supervision of $\mathbf{s}$}

In the second experiment a more challenging task is considered, where the values of $\mathbf{s}$ are partially observed. First, we see that the classification accuracy of the unregularized models is high, this is however coupled with a significant increase in retained sensitive information. Again, the unregularized models seem ill suited for the task of fair classification. Note that the increase in classification accuracy compared to the supervised case is due to the increased weight on $\alpha$. Since there was no baseline from previous work on this task, we performed a more thorough hyperparameter search.

Interestingly, the MI regularizer outperforms the MMD regularizer over the whole spectrum with higher accuracy scores and lower discriminative scores in the partially observed case. It seems the MI regularizer is better suited to handle the estimation uncertainty of $q_\phi(\textbf{s}|\textbf{x})$. Moreover, the MMD requires known $\mathbf{s}$ and since we have less fully supervised training examples, the estimation of the regularizer is worse. Possibly, the MI regularizer is more robust to this problem. 

In conclusion, on the partially supervised task our proposed model outperforms the VFAE with respect to both classification accuracy and sensitive information retained.


\section{Conclusion}

In the paper we proposed the Hierarchical VampPrior Variational Auto-Encoder for learning fair representations. Additionally, we used the mutual information as a regularizer for obtaining fair representations, an alternative to the currently used MMD regularizer. In the experiments we considered two cases: (i) fully observable $\mathbf{s}$ scenario, and (ii) a case with partially supervised $\mathbf{s}$. Especially the second task is interesting because in many real-life situations the information about $\mathbf{s}$ is missing, e.g., in domain adaptation the domain label could be unknown or hard to achieve.

The obtained results on two benchmark datasets show that our model together with the VampPrior obtains very promising results. The MMD regularizer seems to be preferred when the supervised training sample is large enough. Otherwise, using the MI regularizer provides the best results. Nevertheless, more thorough experiments are needed to reach a definite conclusion. Moreover, in this work we used a single Monte Carlo sample to approximate the MI regularizers. The obtained results might possibly be better if a larger sample would be used. We leave investigating these issues for future work.


\section*{Acknowledgements}
The research conducted by Jakub M. Tomczak was funded by the European Commission within the Marie Skłodowska-Curie Individual Fellowship (Grant No. 702666, ''Deep learning and Bayesian inference for medical imaging'').


\end{document}